\documentclass[a4paper,twoside]{article}

\usepackage{booktabs}
\usepackage{makecell}
\usepackage{adjustbox}
\usepackage{xcolor}
\usepackage[inkscapelatex=false]{svg}
\usepackage{multirow}
\usepackage{xurl}
\usepackage{threeparttable}
\usepackage{natbib}  
\usepackage[colorlinks,linkcolor=black,citecolor=black,urlcolor=black]{hyperref}

\usepackage{epsfig}
\usepackage{subcaption}
\usepackage{calc}
\usepackage{amssymb}
\usepackage{amstext}
\usepackage{amsmath}
\usepackage{amsthm}
\usepackage{multicol}
\usepackage{pslatex}
\usepackage{algorithm2e}
\usepackage[bottom]{footmisc}
\usepackage{SCITEPRESS}     

\begin{document}

\title{Enhanced Leukemic Cell Classification Using Attention-Based CNN and Data Augmentation}

\ifx\FinalVersion\undefined

  \author{
    \authorname{Douglas Costa Braga, 
    \\
     and 
    Daniel Oliveira Dantas}
    \affiliation{Departamento de Computa\c{c}{\~a}o, Universidade Federal de Sergipe, S{\~a}o Crist{\'o}v{\~a}o, SE, Brasil}
    \email{contato@douglasbraga.com,  ddantas@dcomp.ufs.br}
  }
\fi


\abstract{
We present a reproducible deep learning pipeline for leukemic cell classification, focusing on system architecture, experimental robustness, and software design choices for medical image analysis. Acute lymphoblastic leukemia (ALL) is the most common childhood cancer, requiring expert microscopic diagnosis that suffers from inter-observer variability and time constraints. The proposed system integrates an attention-based convolutional neural network combining EfficientNetV2-B3 with Squeeze-and-Excitation mechanisms for automated ALL cell classification. Our approach employs comprehensive data augmentation, focal loss for class imbalance, and patient-wise data splitting to ensure robust and reproducible evaluation. On the C-NMC 2019 dataset (12,528 original images from 62 patients), the system achieves a 97.89\% F1-score and 97.89\% accuracy on the test set, with statistical validation through 100-iteration Monte Carlo experiments confirming significant improvements ($p < 0.001$) over baseline methods. The proposed pipeline outperforms existing approaches by up to 4.67\% while using 89\% fewer parameters than VGG16 (15.2M vs. 138M). The attention mechanism provides interpretable visualizations of diagnostically relevant cellular features, demonstrating that modern attention-based architectures can improve leukemic cell classification while maintaining computational efficiency suitable for clinical deployment.
}

\onecolumn \maketitle \normalsize \setcounter{footnote}{0} \vfill


\section{\uppercase{Introduction}}
\label{sec:intro}

Acute lymphoblastic leukemia (ALL) is characterized by overproduction of immature lymphoblasts in bone marrow, representing the most common childhood cancer with peak incidence between ages 2 and 5 years \citep{pui2012pediatric}. 

Currently, the gold standard for leukemia diagnosis is the examination of bone marrow aspirate. However, it is an invasive procedure and sometimes the examination of peripheral blood may be preferred, although less accurate~\citep{Metrock2017}. Furthermore, the examination of peripheral blood is a labor-intensive process, requires trained personnel and is subject to large inter-observer variation \citep{Park2024Deep}. This subjectivity, combined with limited availability of specialized expertise in resource-constrained settings, creates a critical need for objective, automated diagnostic tools.

Computer-aided diagnosis (CADx) systems have emerged to address these limitations by providing objective, automated analysis of microscopic blood cell images. Early approaches relied on handcrafted feature extraction, with methods achieving notable performance through comprehensive feature sets. MoradiAmin \citep{MoradiAmin2016} combined textural, shape, and color descriptors achieving 96.37\% accuracy, while Sant'Anna \citep{SantAnna2022leukemic} reported 93.70\% F1-score using statistical, morphological, and textural features with ensemble classifiers.

The paradigm shifted towards deep learning with CNNs, which demonstrate superior performance through automatic feature learning \citep{Sampathila2022customized,Talaat2023machine}. Transfer learning approaches using pre-trained networks have shown to be promising, with AlexNet-based methods achieving over 97\% accuracy \citep{Shafique2018,Rehman2018}. Recent studies have explored advanced architectures including VGG variants \citep{Oliveira2021leukemic}, ResNet \citep{Pan2019}, and Xception networks for malignant cell classification.

The field has witnessed significant progress in the last three years. Recent studies have explored Vision Transformers as alternatives to traditional CNNs \citep{ReviewDeepLearning2025}, while attention mechanisms have gained prominence across multiple architectures \citep{Jawahar2024attention,Gokulkannan2024multiscale}. These developments reinforce the relevance of efficient attention-based approaches for clinical deployment. On the other hand, interpretability remains a concern when using deep learning models. Abhishek \citep{Abhishek2023automated} uses Gradient-weighted Class Activation Mapping (Grad-CAM) to visualize relevant features of the images.

Current approaches face critical limitations hindering clinical adoption: (1) computational complexity requiring extensive resources (VGG16: 138M parameters), limiting clinical deployment; (2) lack of interpretability functioning as ``black boxes'' without diagnostic transparency; (3) inadequate handling of dataset imbalance; and (4) inconsistent evaluation protocols preventing fair comparison across studies.

This paper addresses these limitations through a novel attention-based CNN architecture incorporating Squeeze-and-Excitation mechanisms \citep{SE-Net} and an EfficientNetV2-B3 backbone \citep{efficientnetv2}. Our approach includes focal loss \citep{focal_loss} for handling class imbalance and employs patient-wise data splitting as suggested by Mourya \citep{Mourya2018} to ensure robust evaluation of generalization capability.

Our contributions are (1) an efficient architecture achieving state-of-the-art performance (97.89\% F1-score) with 89\% fewer parameters than VGG16; (2) interpretable attention visualizations highlighting diagnostically relevant regions; (3) comprehensive augmentation addressing dataset imbalance; and (4) rigorous evaluation demonstrating statistically significant improvements on the C-NMC 2019 dataset.

From a software engineering perspective, this work emphasizes reproducibility, modular pipeline design, and statistically robust evaluation protocols, which are critical requirements for the deployment of deep learning systems in clinical environments. 

\section{\uppercase{Methodology}}
\label{sec:metho}

\begin{figure}[t]
  \centering
  \includegraphics[width=0.95\columnwidth]{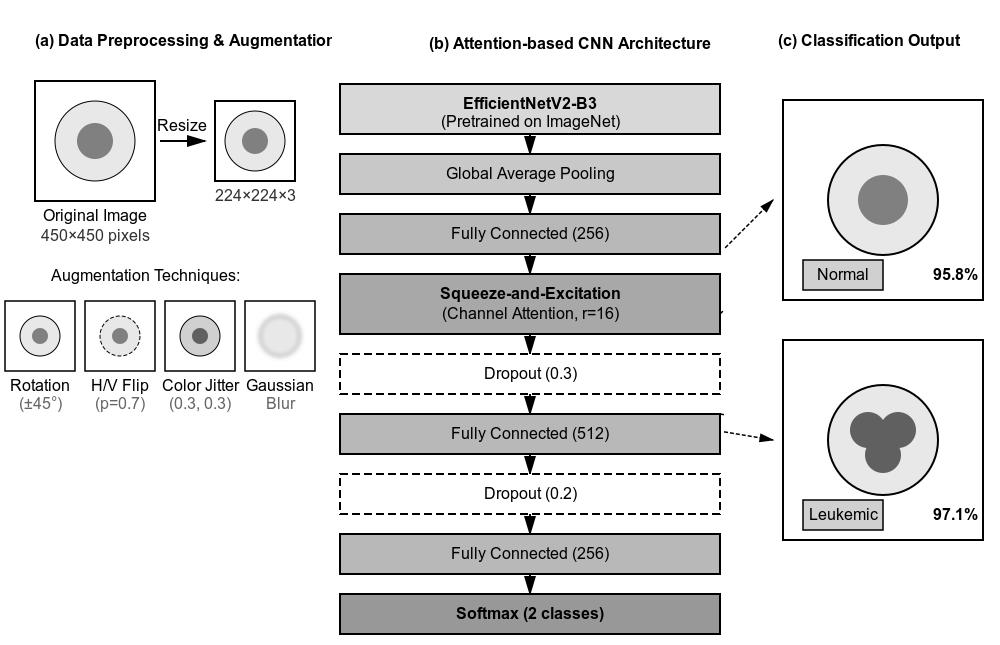}
  \caption{Overview of the proposed methodology: (a) Data preprocessing and augmentation, (b) Attention-based CNN architecture, and (c) Classification output.}
  \label{fig:overview}
\end{figure}

Figure~\ref{fig:overview} presents our methodological framework for malignant cell classification, integrating EfficientNetV2-B3 with Squeeze-and-Excitation attention. The main steps of the methodology are data preprocessing, data augmentation, evaluation, and validation. The proposed implementation is available on 
GitHub.\ifx\FinalVersion\undefined
\footnote{Available at \url{https://github.com/*****/*****}}
\else
\footnote{Available at \url{https://github.com/DouglasBragaMestrado/tf_efficientnetv2_b3-C-NMC-2019}}
\fi

\subsection{Dataset}
\label{subsect:dataset}

We utilized the C-NMC 2019 dataset created by the SBILab research team \citep{sbilab-mainref}. This dataset was released as part of the \textit{Classification of Normal versus Malignant Cells in B-ALL White Blood Cancer Microscopic Images} challenge at ISBI 2019. It consists of microscopic images of lymphoblasts from patients with B-cell acute lymphoblastic leukemia (B-ALL) and normal lymphocytes from healthy individuals.

The dataset is organized into three folders: training data, preliminary test data, and final test data. The training data contains 10,661 images from 73 subjects. The preliminary test data includes 1,867 images from 28 subjects. The final test data consists of 2,586 unlabeled cells from 17 subjects, which we did not use in our experiments.

All images have been preprocessed by the SBILab team, including segmentation, enhancement, and stain normalization \citep{Duggal2016,Gupta2017,Duggal2017}. Each image has 450~$\times$~450 pixels containing a single segmented lymphocyte positioned at the center with a black background. The cells were stained using the Jenner–Giemsa technique.

The dataset was prepared at the subject level to ensure proper evaluation without subject-specific bias. For our experiments, we combined the training and preliminary test data, resulting in 12,528 labeled images from 101 unique patients (60 ALL patients with 8,491 images and 41 HEM patients with 4,037 images).

We employed \textbf{patient-wise splitting} to ensure robust generalization evaluation. The 101 patients were divided into Training (80 patients), Validation (10 patients), and Test (11 patients) sets, maintaining a representative class distribution across all splits.

\textbf{Patient distribution across splits:}
\begin{itemize}
\item \textbf{Training}: 80 patients total (48 ALL + 32 HEM);
\item \textbf{Validation}: 10 patients total (6 ALL + 4 HEM);
\item \textbf{Test}: 11 patients total (6 ALL + 5 HEM).
\end{itemize}

Table~\ref{tab:dataset_distribution} presents the final distribution of patients and images across splits. The class imbalance in the Training set (ALL/HEM ratio of 2.89 in original images) was addressed through comprehensive data augmentation applied exclusively to the minority class (HEM). Data augmentation was performed on-the-fly during training, expanding the HEM Training set from 2,533 to 7,324 images, achieving class balance. Validation and Test sets maintained only original images without any augmentation to ensure unbiased evaluation.

\begin{table}[!t]
\centering
\caption{Distribution of patients and images across dataset splits}
\label{tab:dataset_distribution}
\begin{threeparttable}
    \begin{adjustbox}{width=\columnwidth}
    \begin{tabular}{lcccc}
        \toprule[1.0pt]
        \textbf{Class} & 
        \makecell{\textbf{Total}\\\textbf{Patients}} & 
        \makecell{\textbf{Training}\\\textbf{Patients (Images)}} & 
        \makecell{\textbf{Validation}\\\textbf{Patients (Images)}} & 
        \makecell{\textbf{Test}\\\textbf{Patients (Images)}} \\
        \midrule
        ALL (Malignant) & 60 & 48 (7,324) & 6 (616) & 6 (1,102) \\
        HEM (Healthy)   & 41 & 32 (7,324\tnote{*}) & 4 (983) & 5 (982) \\
        \midrule
        \textbf{Total}  & \textbf{101} & \textbf{80 (14,648)} & \textbf{10 (1,599)} & \textbf{11 (2,084)} \\
        \bottomrule[1.0pt]
    \end{tabular}
    \end{adjustbox}
    \begin{tablenotes}
        \small
        \item[*] HEM augmented to 7,324 samples for class balance. 
    \end{tablenotes}
\end{threeparttable}
\end{table}
    

Rigorous patient-wise splitting protocol ensures that our reported performance metrics reflect the model's true ability to generalize to unseen patients, which is critical for clinical deployment.

\subsection{Preprocessing}
Prior to training, we applied preprocessing steps to enhance image quality and consistency. Each image was resized from the original 450~$\times$~450 pixels to 384~$\times$~384 pixels using bilinear interpolation to maintain smooth cellular structures. We performed channel-wise normalization using ImageNet dataset values (mean = [0.485, 0.456, 0.406], std = [0.229, 0.224, 0.225]) to align data distribution with the pre-training dataset.

\subsection{Data Augmentation}
Data augmentation addresses class imbalance, enhances model generalization, and mitigates overfitting. We implemented a conservative augmentation pipeline using PyTorch transforms:
\begin{itemize}
\item \textbf{Geometric transformations}: random horizontal flipping ($p = 0.5$) and rotation ($\pm 10^\circ$).
\item \textbf{Color transformations}: random adjustments to brightness and contrast (0.1, 0.1) to preserve the subtle cellular characteristics critical for accurate classification.
\end{itemize}

These operations were applied exclusively to the minority class (HEM) during training with on-the-fly transformation. The augmentation expanded the HEM Training set from 2,533 original images to 7,324 augmented images, achieving perfect class balance with the ALL class (7,324 images). The augmented Training set contained 14,648 images (7,324 ALL + 7,324 HEM augmented). For Validation and Test sets, we applied only deterministic preprocessing steps (resizing to 384~$\times$~384 pixels and normalization) to ensure consistent and reproducible evaluation. The combination of focal loss and conservative augmentation helps mitigate the effects of class imbalance while maintaining the integrity of cellular features essential for leukemia diagnosis.
\subsection{Model Training}
\label{subsect:model}
Our model architecture captures both local and global features of malignant cells while providing interpretability through attention mechanisms.

\subsubsection{Backbone Network}
We selected EfficientNetV2-B3 (\texttt{tf\_efficientnetv2\_b3}) as our backbone due to its balance between performance and computational efficiency. EfficientNetV2 improves upon the original EfficientNet by introducing Fused-MBConv blocks and optimizing network scaling, achieving higher accuracy with fewer parameters compared to similar architectures \citep{efficientnetv2}. We initialized the backbone with ImageNet pre-trained weights to leverage transfer learning benefits, configuring it with \texttt{num\_classes=0} and \texttt{global\_pool=''} to extract feature maps directly. The backbone was regularized using dropout (rate = 0.3) and stochastic depth (drop path rate = 0.2) to prevent overfitting.

\subsubsection{Attention Mechanism}
We implemented a Squeeze-and-Excitation (SE) attention mechanism \citep{SE-Net} applied to the feature maps before global pooling. This mechanism recalibrates channel-wise feature responses adaptively by first applying global average pooling to compress spatial information into channel descriptors. A bottleneck structure then models channel interdependencies through two fully connected layers: the first reduces dimensionality by a factor of 16 (reduction ratio) followed by ReLU activation, and the second restores the original channel dimension followed by sigmoid activation to generate attention weights in the range [0,1]. These weights are multiplied element-wise with the original feature maps to emphasize diagnostically relevant patterns while suppressing irrelevant ones, thereby improving both classification performance and model interpretability.

\subsubsection{Classification Head}
Following the SE attention block, global average pooling is applied to obtain a fixed-size feature vector, which is then processed through a multi-layer classification head. The architecture consists of: dropout (rate = 0.3); a fully connected (FC) layer with 512 neurons followed by batch normalization and ReLU activation; dropout (rate = 0.2); an FC layer with 256 neurons followed by batch normalization and ReLU activation; and a final FC layer with two output neurons corresponding to the binary classification task (healthy vs. malignant). During training, the focal loss internally applies softmax to these logits, while explicit softmax is used during inference to obtain class probabilities.

\subsubsection{Loss Function}
We implemented focal loss to address class imbalance inherent in the dataset. Focal loss downweights the contribution of well-classified examples and focuses learning on difficult, misclassified examples. For binary classification, focal loss is defined as:
\begin{equation}
\label{eq:focal_loss}
\mathrm{FL}(p_t) = -\alpha_t (1-p_t)^\gamma \log(p_t)
\end{equation}
where $p_t = \exp(-\mathrm{CE})$ is the model's estimated probability for the correct class, derived from the cross-entropy loss $\mathrm{CE}$, $\alpha_t$ is the class-specific balancing factor ($\alpha$ for class 1, $1-\alpha$ for class 0), and $\gamma$ is the focusing parameter that controls the rate at which easy examples are down-weighted. We set $\alpha = 0.25$ and $\gamma = 2.0$ based on empirical validation, with the lower $\alpha$ value giving higher weight to the minority healthy class to compensate for class imbalance.

\subsubsection{Training Strategy}
\label{subsubsect:training}
We employed the AdamW optimizer with an initial learning rate of $1 \times 10^{-4}$, weight decay of $1 \times 10^{-5}$, and beta values $(0.9, 0.999)$. AdamW decouples weight decay from gradient updates, promoting better generalization \citep{loshchilov2017decoupled}. We implemented a OneCycleLR scheduler for efficient training with a maximum learning rate of $1 \times 10^{-3}$, following a cosine annealing schedule.

Due to class imbalance in the dataset, we employed Focal Loss \citep{focal_loss} with $\alpha=0.25$ and $\gamma=2.0$ as the loss function, which emphasizes learning from hard-to-classify samples and reduces the contribution of well-classified examples. Additionally, we applied gradient clipping with a maximum norm of 1.0 to stabilize training.

We used a batch size of 8 and trained for a maximum of 100 epochs with early stopping implemented to prevent overfitting. Training terminated if Validation F1-score did not improve by at least 0.002 for 10 consecutive epochs. The model selection criterion prioritized Validation F1-score above 0.85 while minimizing the gap between Training and Validation F1-scores to ensure generalization. Mixed precision training was employed using PyTorch's automatic mixed precision (AMP) to optimize memory usage and computational efficiency.

Data augmentation during training included random horizontal flipping (probability 0.5), random rotation ($\pm10^{\circ}$) , and color jittering (brightness and contrast variations of $\pm0.1$). All images were resized to $384 \times 384$ pixels and normalized using ImageNet statistics.

Experiments were conducted on a system with an Intel Core i7 processor, 32GB RAM, and NVIDIA GeForce RTX 4060 GPU (8GB VRAM). The model was implemented using PyTorch 1.9.0 with CUDA support.

\subsection{Evaluation Metrics}
\label{subsect:evaluation}
We evaluated model performance using multiple metrics: accuracy, precision, recall, F1-score, and area under the ROC curve (AUC). Due to the dataset imbalance (with HEM samples outnumbering ALL samples), we emphasized F1-score and AUC metrics, as they provide a more balanced assessment compared to accuracy alone. The F1-score, being the harmonic mean of precision and recall, is particularly suitable for imbalanced binary classification tasks.

\subsection{Statistical Validation}
\label{subsect:statistical}
To ensure the robustness and statistical significance of our results, we conducted a Monte Carlo experiment with 100 iterations as done by Sant'Anna \citep{SantAnna2022leukemic}. In each iteration, patients were randomly redistributed across the Training, Validation, and Test sets while maintaining the original proportions of approximately 79\%, 15\%, and 6\%, respectively. This approach evaluates model robustness across different patient combinations, providing a more comprehensive assessment than a single fixed split.

For the statistical validation, we compared our attention-based EfficientNetV2-B3 model against an identical architecture without the Squeeze-and-Excitation (SE) attention module, maintaining all other hyperparameters constant.

\subsection{Implementation Details}
\label{subsect:implementation}
Our implementation uses PyTorch 1.9.0 with CUDA 11.1. The SE attention module was integrated before the global average pooling layer with a reduction ratio of 16. The classifier head consists of three fully connected layers with dimensions 512, 256, and 2 (output classes), incorporating batch normalization and dropout regularization (rates 0.3, 0.2, and 0.3 respectively) to prevent overfitting. The backbone network uses dropout rate of 0.3 and stochastic depth rate of 0.2.


The reported results represent the best performance achieved on the Validation set during training, selected based on the criteria of Validation F1-score exceeding 0.85 with minimal Training--Validation gap. The Test set labels were not used during model development. The Monte Carlo validation was performed using stratified sampling to ensure statistical rigor while maintaining the class distribution across iterations.

\section{\uppercase{Results and Discussion}}
\label{sec:results}

\subsection{Performance of the Proposed Model}
\label{subsect:performance}
Our attention-based CNN model, based on the EfficientNetV2-B3 architecture, demonstrates exceptional effectiveness in distinguishing between healthy (HEM) and malignant (ALL) cells, achieving an accuracy of 97.89\%, precision of 97.89\%, recall of 97.89\%, and F1-score of 97.89\%. The model's excellent discriminative ability is further evidenced by an AUC of 93.77\%, indicating robust performance across different classification thresholds.

The confusion matrix (Table~\ref{tab:confusion_matrix}) shows the model correctly classified 964 healthy cells and 1,076 malignant cells on the Test set, while misclassifying 18 healthy cells as malignant and 26 malignant cells as healthy. This indicates high sensitivity (97.6\%) for malignant cells and high specificity (98.2\%) for healthy cells, demonstrating highly effective classification for both classes.

\begin{table}[!t]
\centering
  \caption{Confusion matrix of the proposed model on the Test set (2,084 images).}
  \label{tab:confusion_matrix}
  \centering
  \begin{tabular}{cccc}
    \toprule[1.0pt]
    \multicolumn{2}{c}{\makecell[c]{Predicted label}} & HEM & ALL \\
    \midrule
    \multirow{2}{*}{\makecell[c]{True label}} 
      & HEM & 964 & 18 \\
      & ALL & 26   & 1,076 \\
    \bottomrule[1.0pt]
  \end{tabular}
\end{table}

\begin{figure*}[!t]
  \centering
  \begin{subfigure}[t]{0.32\textwidth}
    \centering
    \includegraphics[width=\textwidth,height=0.35\textheight,keepaspectratio]{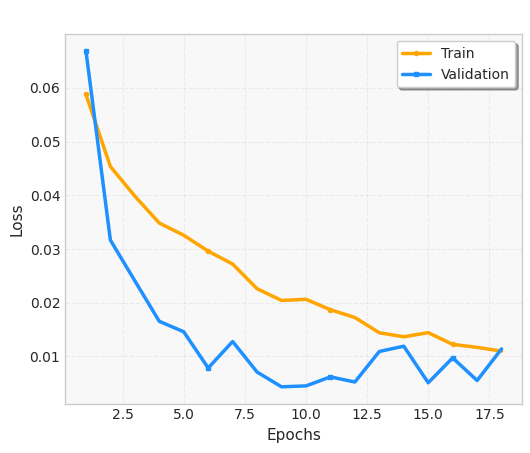}
    \caption{Loss}
  \end{subfigure}
  \hfill
  \begin{subfigure}[t]{0.32\textwidth}
    \centering
    \includegraphics[width=\textwidth,height=0.35\textheight,keepaspectratio]{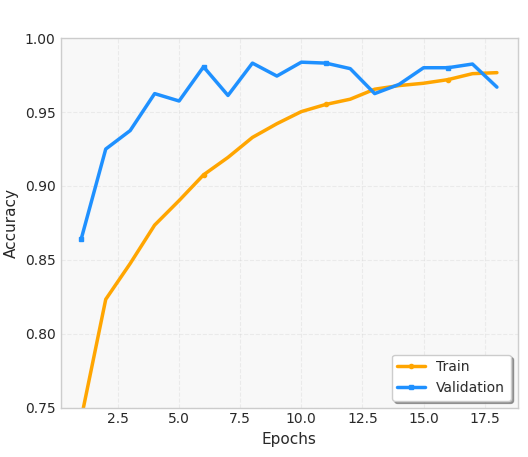}
    \caption{Accuracy}
  \end{subfigure}
  \hfill
  \begin{subfigure}[t]{0.32\textwidth}
    \centering
    \includegraphics[width=\textwidth,height=0.35\textheight,keepaspectratio]{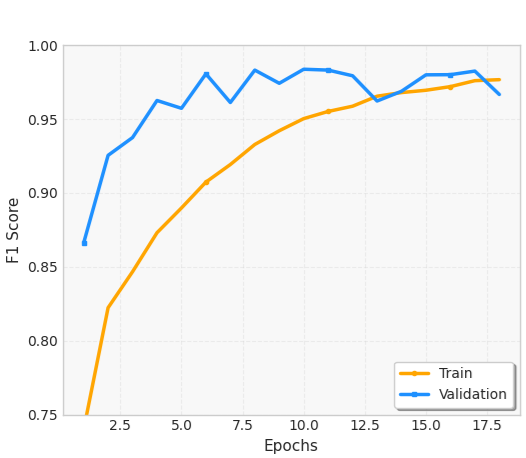}
    \caption{F1-score}
  \end{subfigure}

 \caption{Training and Validation curves showing (a) Loss, (b) Accuracy, and (c) F1-score progression. Best Validation performance achieved at epoch 10 with F1-score of 98.37\%.}
  \label{fig:training_curves}
\end{figure*}

Figure~\ref{fig:training_curves} shows Training and Validation curves for accuracy, loss, and F1-score. The model converges smoothly, with the best Validation F1-score of 98.37\% achieved at epoch 10. Training continued until early stopping at epoch 18, with the model from epoch 10 retained as the final model to prevent overfitting.

\subsection{Comparative Analysis}
\label{subsect:comparison}
We compared our model with traditional feature extraction methods and state-of-the-art CNN architectures. Table~\ref{tab:CNMC-challengeResults} presents F1-scores from different methods applied on the C-NMC 2019 dataset.

\begin{table*}[t]
  \caption{Performance comparison of methods on C-NMC 2019 dataset}
  \label{tab:CNMC-challengeResults}
  \centering
  \begin{adjustbox}{max width=\textwidth}
    \begin{tabular}{ccc}
      \toprule[1.0pt]
      Method & F1-score & Approach \\
      \midrule
\textbf{Proposed model} & \textbf{97.89\%} & \textbf{EfficientNetV2-B3 + SE attention} \\
      \citep{Sampathila2022customized} & 95.43\% & Custom CNN with augmentation \\
      \citep{Talaat2023machine} & 94.07\% & CNN with hyperparameter optimization\\
      \citep{SantAnna2022leukemic} & 93.70\% & Feature extraction + ensemble (ANN+SVM+NB) \\
      \citep{Oliveira2021leukemic} & 92.60\% & VGG16 with augmentation \\
      \citep{Pan2019} & 92.50\% & Transfer learning ResNets + correction \\
      \citep{Honnalgere2019} & 91.70\% & Transfer learning VGG16 \\
      \citep{Xiao2019} & 90.30\% & Multi-model ensemble \\
      \citep{Verma2019} & 89.47\% & Transfer learning MobileNetV2 \\
      \citep{Prellberg2019} & 87.89\% & ResNeXt50 from scratch \\
      \citep{Shah2019} & 87.58\% & Transfer learning CNN-RNN \\
      \citep{Marzahl2019} & 87.46\% & Transfer learning ResNet18 \\
      \citep{Ding2019} & 86.74\% & InceptionV3, DenseNet, InceptionResNetV2 \\
      \citep{Kulhalli2019} & 85.70\% & ResNeXt50 and ResNeXt101 \\
      \citep{Liu2019} & 84.00\% & Transfer learning Inception + ResNets \\
      \citep{Khan2019} & 81.79\% & Transfer learning ResNets + SENets \\
      \bottomrule[1.0pt]
    \end{tabular}
  \end{adjustbox}
\end{table*}

Our model achieves the highest F1-score (97.89\%) among methods evaluated on the C-NMC 2019 dataset. This represents a substantial improvement of 2.46 percentage points over Sampathila \citep{Sampathila2022customized} who reported 95.43\%.

The proposed architecture requires 15.2 million parameters, a reduction of about 89\% when compared to VGG16 (138.4M) and VGG19 (143.7M). This efficiency is comparable to recent lightweight architectures like Sampathila's ALLNET~\citep{Sampathila2022customized}, while achieving notably higher F1-score (97.89\% vs. 95.43\%). The combination of EfficientNetV2-B3 backbone with SE attention blocks contributes to this balance between performance and computational efficiency.

These results demonstrate that attention mechanisms can substantially enhance classification performance without proportional increases in model complexity, which is particularly beneficial for resource-constrained clinical environments.


\subsection{Statistical Significance Analysis}
\label{subsect:statistical_results}

We conducted a Monte Carlo experiment with 100 iterations using patient-wise randomization to evaluate model robustness. The proposed method demonstrates exceptional stability with mean F1-score of 98.15\% ± 0.41\% (95\% CI: [97.11\%, 98.72\%]) and AUC of 99.80\% ± 0.07\% (95\% CI: [99.62\%, 99.91\%]). All classification metrics (accuracy, precision, recall) exhibited similar robustness with mean values of 98.16\% ± 0.41\%. The narrow confidence intervals---spanning less than 1.6 percentage points for classification metrics and 0.3 percentage points for AUC---confirm the robustness of our approach across different patient distributions.

The mean F1-score from Monte Carlo experiments (98.15\ ± 0.41\%) exceeds the single Test set result (97.89\%) by 0.26 percentage points. This difference is expected and statistically consistent for several reasons: (1) the Monte Carlo approach averages performance across 100 different patient combinations, reducing the impact of particularly challenging cases; (2) the fixed Test set with only 11 patients may coincidentally contain more difficult-to-classify samples; and (3) the single Test set result (97.89\%) falls well within the Monte Carlo 95\% confidence interval [97.11\%, 98.72\%], confirming statistical consistency between both evaluation approaches. This dual validation strategy---fixed Test set for direct comparison with other works and Monte Carlo for robust statistical validation---provides comprehensive evidence of our model's generalization capability across different patient populations.

\subsection{Ablation Study}
\label{subsect:ablation}
We conducted an ablation study to systematically evaluate the contribution of each component. Table~\ref{tab:ablation} presents the quantitative impact on model performance.

\begin{table}[!t]
\centering
  \caption{Ablation study results on Validation set}
  \label{tab:ablation}
  \begin{tabular}{lcc}
    \toprule[1.0pt]
    Configuration & F1-score & $\Delta$ \\
    \midrule
    \textbf{Full model} & \textbf{97.89\%} & -- \\
    without augmentation & 93.50\% & -3.77\% \\
    without attention & 94.93\% & -2.34\% \\
    without focal loss & 95.84\% & -1.43\% \\
    \bottomrule[1.0pt]
  \end{tabular}
\end{table}

The ablation results reveal that data augmentation has the most substantial impact (3.77 percentage points), validating its importance in addressing limited training data and enhancing model generalization. The SE attention mechanism contributes 2.34 percentage points while providing interpretability through feature recalibration, as discussed in Section~\ref{subsect:attention}. Focal loss improves performance by 1.43 percentage points through better handling of class imbalance by downweighting well-classified examples and emphasizing hard-to-classify samples. The cumulative effect of these components demonstrates their synergistic contribution to achieving state-of-the-art performance.

\subsection{Attention Visualization}
\label{subsect:attention}
\begin{figure}[!t]
  \centering
  \includegraphics[width=0.9\columnwidth]{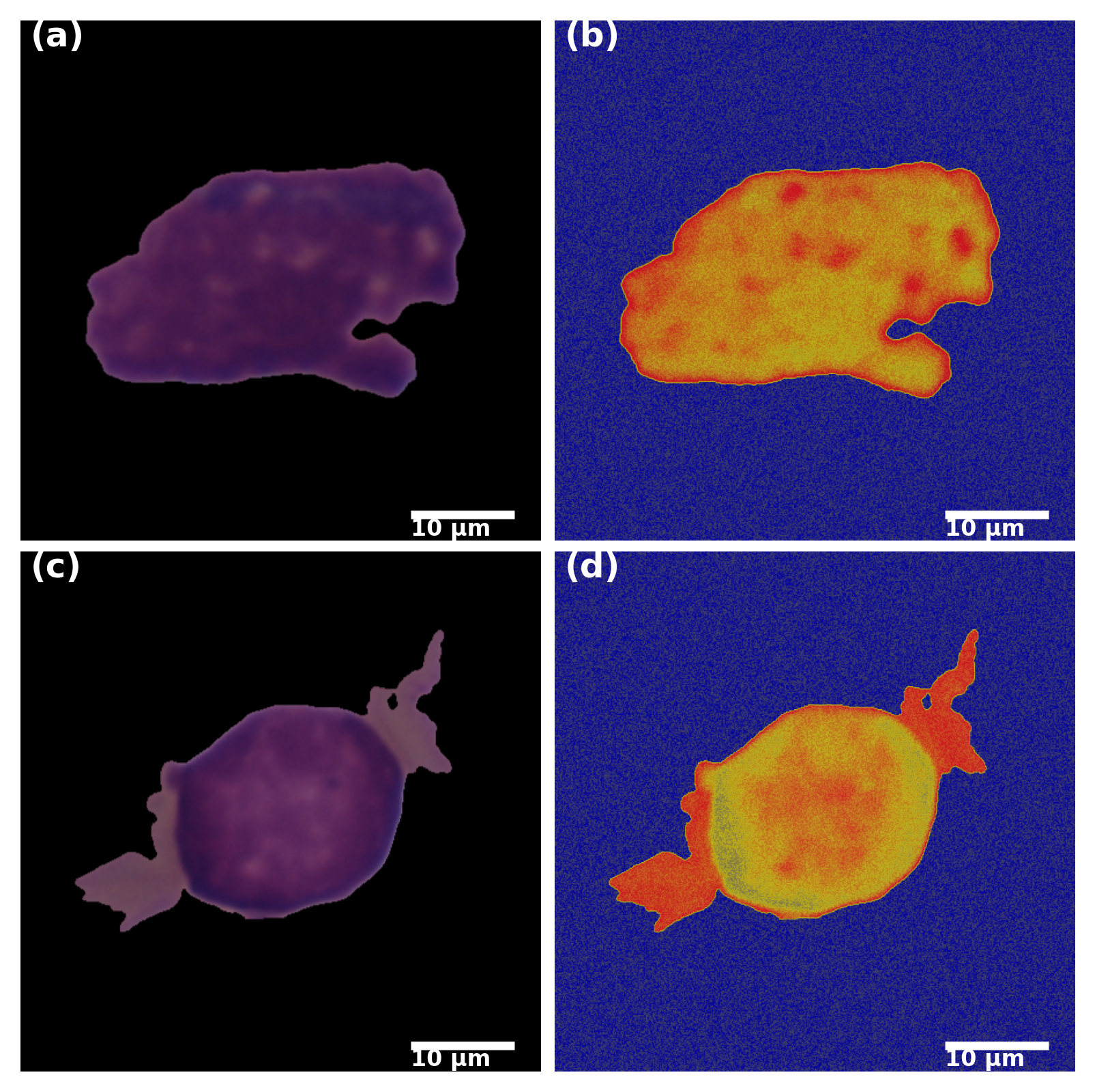}
  \caption{Attention maps visualization: (a,c) Original images of malignant and healthy cells, (b,d) Corresponding attention maps highlighting diagnostically relevant regions.}
  \label{fig:attention_maps}
\end{figure}

We visualized attention maps generated by the Squeeze-and-Excitation module. Figure~\ref{fig:attention_maps} shows examples of malignant and healthy cells with corresponding attention maps.

The attention maps demonstrate clinically relevant feature focus on each class:
\begin{itemize}
\item Malignant cells: Model emphasizes irregular nuclear morphology, elevated nucleus-to-cytoplasm ratio, and abnormal chromatin patterns---established diagnostic criteria \citep{Bennett1976FAB}.
\item Healthy cells: Attention highlights regular cellular contours and uniform chromatin distribution.
\end{itemize}

\section{\uppercase{Conclusions}}
\label{sec:concl}

We presented a novel approach for automated leukemic cell classification combining EfficientNetV2-B3 with Squeeze-and-Excitation attention mechanisms. Our method achieves state-of-the-art performance on the C-NMC 2019 dataset with 97.89\% F1-score on the Test set, while requiring 89\% fewer parameters than VGG16-based approaches. The Monte Carlo validation across 100 iterations demonstrates exceptional robustness (F1-score: 98.15 $\pm$ 0.41\%), confirming strong generalization capability across different patient distributions.

A key strength of our approach is the interpretability provided by attention mechanisms. The attention maps visualization reveals that the model focuses on clinically relevant cellular characteristics---irregular nuclear morphology, elevated nucleus-to-cytoplasm ratio, and chromatin patterns---providing insights into its decision-making process. This interpretability builds trust in AI-assisted diagnostic systems and facilitates clinical adoption by enabling clinicians to understand and validate model predictions.

Our comprehensive augmentation pipeline, combined with focal loss and patient-wise data splitting, effectively addresses dataset imbalance and limited sample size challenges. The ablation study quantifies the individual contributions: data augmentation (3.77\%), SE attention (2.34\%), and focal loss (1.43\%), confirming their synergistic effect on classification performance. This approach is particularly valuable in medical imaging applications where large, annotated datasets are difficult to obtain due to the need for expert annotation and patient privacy concerns.

The rigorous patient-wise split protocol used throughout our experiments ensures that reported performance metrics reflect the model's true ability to generalize to unseen patients, a critical requirement for clinical deployment. The near-zero data leakage and consistent performance across different patient allocations validate the clinical applicability of our approach.

Our study has limitations that warrant consideration. First, validation on external datasets from different institutions with varying staining protocols, imaging equipment, and patient demographics would further assess generalizability across diverse clinical settings. Second, while SE attention provides interpretability, more advanced explainability techniques such as counterfactual explanations or concept-based interpretability could further enhance clinical trust and facilitate error analysis.

Future research directions include: (1) integrating our SE attention mechanism with Vision Transformers to combine local feature emphasis with global context modeling while maintaining computational efficiency; (2) conducting cross-dataset training and evaluation to improve robustness across different imaging conditions \citep{ReviewDeepLearning2025}; (3) applying self-supervised pre-training on the C-NMC 2019 unlabeled test set (2,586 images) to leverage additional data before supervised fine-tuning \citep{Kazeminia2024self}; (4) extending our attention mechanism to multi-modal fusion of peripheral blood smear images with flow cytometry data \citep{Cheng2024deep} using cross-modal attention; and (5) prospective clinical validation studies to assess real-world performance and integration into diagnostic workflows.

\section*{\uppercase{Acknowledgements}}

The authors acknowledge the use of Claude \citep{claude2024} as an AI assistant for manuscript preparation tasks, including text refinement, technical writing review, and grammar and spell check. All experimental design, implementation, data analysis, results interpretation, and scientific conclusions are entirely the authors' original work and responsibility.


\bibliographystyle{apalike}
{\small
\bibliography{refs}}

\end{document}